# Relating Turing's Formula and Zipf's Law

**Christer Samuelsson**
Universität des Saarlandes, FR 8.7, Computerlinguistik
Postfach 1150, D-66041 Saarbrücken, Germany
christer@coli.uni-sb.de

## Abstract

An asymptote is derived from Turing's local reestimation formula for population frequencies, and a local reestimation formula is derived from Zipf's law for the asymptotic behavior of population frequencies. The two are shown to be qualitatively different asymptotically, but nevertheless to be instances of a common class of reestimation-formula-asymptote pairs, in which they constitute the upper and lower bounds of the convergence region of the cumulative of the frequency function, as rank tends to infinity. The results demonstrate that Turing's formula is qualitatively different from the various extensions to Zipf's law, and suggest that it smooths the frequency estimates towards a geometric distribution.

## 1 Introduction

Turing's formula [Good 1953] and Zipf's law [Zipf 1935] indicate how population frequencies in general tend to behave. Turing's formula estimates locally what the frequency count of a species that occurred $x$ times in a sample really would have been, had the sample accurately reflected the underlying population distribution. Zipf's law prescribes the asymptotic behavior of the relative frequencies of species as a function of their rank. The ranking scheme in question orders the species by frequency, with the most common species ranked first. The reason that these formulas are of interest in computational linguistics is that they can be used to improve probability estimates from relative frequencies, and to predict the frequencies of unseen phenomena, e.g., the frequency of previously unseen words encountered in running text.

Due to limitations in the amount of available training data, the so-called sparse-data problem, estimating probabilities directly from observed relative frequencies may not always be very accurate. For this reason, Turing's formula, in the incarnation of Katz's back-off scheme [Katz 1987], has become a standard technique for improving parameter estimates for probabilistic language models used by speech recognizers. A more theoretical treatment of Turing's formula itself can be found in [Nádas 1985].

Zipf's law is commonly regarded as an empirically accurate description of a wide variety of (linguistic) phenomena, but too general to be of any direct use. For a bit of historic controversy on Zipf's law, we refer to [Simon 1955], [Mandelbrot 1959], and subsequent articles in *Information and Control*. The model presented there for the stochastic source generating the various Zipfian distributions is however linguistically highly dubious: a version of the monkey-with-typewriter scenario.

The remainder if this article is organized as follows. In Section 2, we induce a recurrence equation from Turing's local reestimation formula and from this derive the asymptotic behavior of the relative frequency as a function of rank, using a continuum approximation. The resulting probability distribution is then examined, and we rederive the recurrence equation from it. In Section 3, we start with the asymptotic behavior stipulated by Zipf's law, and derive a recurrence equation similar to that associated with Turing's formula, and from this induce a corresponding reestimation formula. We then rederive the Zipfian asymptote from the established recurrence equation. In Section 4, similar techniques are used to establish the asymptotic behavior inherent in a general class of recurrence equations, parameterized by a real-valued parameter, and then to rederive the recurrence equations from their asymptotes. The convergence region of this parameter for the cumulative of the frequency function, as rank approaches infinity, is also investigated. In Section 5, we summarize the results, discuss how they might be used practically, and compare them with related work.

## 2 An Asymptote for Turing's Formula

Turing's formula reestimates population frequencies locally:

$$x^* = (x+1) \cdot \frac{N_{x+1}}{N_x} \qquad (1)$$

Here $N_x$ is the number of species with frequency count $x$, and $x^*$ is the improved estimate of $x$. Let $N$ be the size of the entire population and note that

$$N = \sum_{x=1}^{X} x \cdot N_x \quad \text{and} \quad f_x = \frac{x}{N}$$

where $X$ is the count of the most populous species and $f_x$ is the relative frequency of any species with frequency count $x$.

Let $r(x)$ be the rank of the last species with frequency count $x$, where the most frequent species is ranked first. This means that quite in general

$$r(x) = \sum_{k=x}^{X} N_k$$

$$N_x = \sum_{k=x}^{X} N_k - \sum_{k=x+1}^{X} N_k = r(x) - r(x+1) \qquad (2)$$

### 2.1 A continuum approximation

We first make a continuum approximation by extending $N_x$ from the integer points $x = 0, 1, 2, \ldots, X$ to a continuous function $N(x)$ on $[0, \infty)$. This means that

$$r(x) = \sum_{k=x}^{X} N_k \approx \int_{x}^{X} N(y)\, dy$$

Differentiating this w.r.t. $x$, the lower bound of the integral, yields

$$\frac{dr(x)}{dx} = \frac{d}{dx} \int_{x}^{X} N(y)\, dy = -N(x)$$

and using the chain rule for differentiation yields

$$\frac{dr}{df} = \frac{dr}{dx} \cdot \frac{dx}{df} = -N(x) \cdot N \qquad (3)$$

Continuum approximations are useful techniques for establishing the dependence of a sum on its bounds, to the leading term, and for determining convergence. For example, if we wish to study the sum $\sum_{k=1}^{n} k^2$, we note that the corresponding integral $\int_{1}^{n} x^2 \, dx = \frac{n^3 - 1}{3}$ and conclude that the sum behaves like $n^3$. The exact formula is $\frac{2n^3 + 3n^2 + n}{6}$, so we in fact even got the leading coefficient right. Likewise, we can establish for what values of $\alpha$ the sum $\sum_{k=1}^{\infty} k^\alpha$ converges by explicitly calculating

$$\int_{1}^{\infty} x^\alpha \, dx = \begin{cases} [\ln x]_1^\infty & \text{for } \alpha = -1 \\ \left[\dfrac{x^{\alpha+1}}{\alpha + 1}\right]_1^\infty & \text{for } \alpha \neq -1 \end{cases}$$

indicating that the integral, and thus the sum, converge for $\alpha < -1$ and diverge for $\alpha \geq -1$.

We have to be a bit careful with the transition to the continuous case. We will first let $N$ become large and then establish what happens for small, but non-zero, values of $f = \frac{x}{N}$. So although $x$ will be small compared to $N$, it will be large compared to any constant $C$. This means that

$$f = \lim_{N \to \infty} \frac{x}{N} = \lim_{N \to \infty} \frac{x + C}{N}$$

for any additive constant $C$, and we may approximate $x + C$ with $x$, motivating $\frac{1}{x + 1} \approx \frac{1}{x}$ and similar approximations in the following.

## 2.2 The asymptotic distribution

For an ideal Turing population, we would have $x = x^*$. This gives us the recurrence equation

$$N_{x+1} = \frac{x}{x + 1} \cdot N_x \tag{4}$$

implying that there are equally many inhabitants for frequency count $x$ as for frequency count $x + 1$. This introduces several additional constraints, namely

$$x \cdot N_x = 1 \cdot N_1 \quad \text{and thus} \quad N_x = \frac{N_1}{x} \tag{5}$$

$$N = X \cdot N_1 \quad \text{and thus} \quad f_X = \frac{X}{N} = \frac{1}{N_1}$$

We are now prepared to derive the asymptotic behavior of the relative frequency $f(r)$ of species as a function of their rank $r$ implicit in Eq. (4). Combining Eq. (5) with Eq. (3) yields

$$\frac{dr}{df} = -N(x) \cdot N = -\frac{N_1}{x} \cdot N = -\frac{N_1}{f}$$

This determines the rank $r(f)$ as a function of the relative frequency $f$:

$$r(f) = C - N_1 \ln f \tag{6}$$

Inverting this gives us the sought-for function $f(r)$:

$$f(r) = e^{\frac{C-r}{N_1}} = C' e^{-\frac{r}{N_1}}$$

Utilizing the fact that the relative frequencies should be normalized to one, we find that

$$1 \;=\; \int_1^\infty f(r)\,dr \;=\; C' \cdot N_1 e^{-\frac{1}{N_1}}$$

and that thus "Turing's asymptotic law" is

$$f(r) \;=\; \frac{1}{N_1} e^{-\frac{r-1}{N_1}} \tag{7}$$

Note that, reassuringly, the relative frequency of the most populous species, $f_X$, is preserved:

$$f(1) \;=\; \frac{1}{N_1} \;=\; \frac{X}{N} \;=\; f_X$$

Upon examining the frequency function $\frac{1}{N_1} e^{-\frac{r-1}{N_1}}$, we realize that we have an exponential distribution with intensity parameter $\frac{1}{N_1}$, the probability of the most common species. This distribution was created by approximating our original discrete distribution with a continuous one. The discrete counterpart of an exponential distribution is a geometric distribution

$$P(r) \;=\; p \cdot (1-p)^{r-1} \qquad r = 1, 2, \ldots$$

parameterized by $p$, the probability of some outcome occurring in one trial. $P(r)$ can then be interpreted as the probability of waiting $r$ trials for the first occurrence of the outcome. Thus, Turing's formula seems to be smoothing the frequency estimates towards a geometric distribution.

## 2.3 Rederiving Turing's formula

To test our derivation of the asymptotic equation (7) from the recurrence equation (4), we will attempt to rederive Eq. (4) from Eq. (7). Since Eq. (7) implies Eq. (6), we start from the latter and establish that

$$r(x) \;=\; C - N_1 \ln \frac{x}{N}$$

Inserting this into Eq. (2) yields

$$N_x \;=\; r(x) - r(x+1) \;=\; -N_1 \ln \frac{x}{N} + N_1 \ln \frac{x+1}{N} \;=\; N_1 \ln \frac{x+1}{x} \;=\; N_1 \ln(1 + \frac{1}{x})$$

This means that

$$\left| \frac{N_{x+1}}{N_x} - \frac{x}{x+1} \right| \;=\; \left| \frac{\ln\left(1 + \frac{1}{x+1}\right)}{\ln\left(1 + \frac{1}{x}\right)} - \frac{x}{x+1} \right| \;=\; \left| \frac{(x+1)\ln\left(1 + \frac{1}{x+1}\right) - x \ln\left(1 + \frac{1}{x}\right)}{(x+1)\ln\left(1 + \frac{1}{x}\right)} \right|$$

We first note that

$$\frac{1}{x+1} \;\leq\; \ln\left(1 + \frac{1}{x}\right) \;\leq\; \frac{1}{x}$$

We also note that the numerator can be written as $g(x+1) - g(x)$ for $g(y) = y \ln\left(1 + \frac{1}{y}\right)$, which in turn can be written as $\int_x^{x+1} \frac{d}{dy} g(y)\,dy$, i.e., as $\int_x^{x+1} \left( \ln\left(1 + \frac{1}{y}\right) - \frac{1}{y+1} \right) dy$. We further note that if $A \leq h(y) \leq B$ on $(a, b)$, then $A(b-a) \leq \int_a^b h(y)\, dy \leq B(b-a)$. Hence

$$0 \leq \frac{(x+1)\ln\left(1+\frac{1}{x+1}\right) - x\ln\left(1+\frac{1}{x}\right)}{(x+1)\ln\left(1+\frac{1}{x}\right)} = \frac{\int_x^{x+1}\left(\ln\left(1+\frac{1}{y}\right) - \frac{1}{y+1}\right)dy}{(x+1)\ln\left(1+\frac{1}{x}\right)} \leq$$

$$\leq \int_x^{x+1}\left(\frac{1}{y} - \frac{1}{y+1}\right)dy = \int_x^{x+1}\frac{1}{y(y+1)}\,dy \leq \int_x^{x+1}\frac{1}{y^2}\,dy \leq \frac{1}{x^2}$$

We have thus proved that $\left|\frac{N_{x+1}}{N_x} - \frac{x}{x+1}\right| \leq \frac{1}{x^2}$ and since we assume that $x \gg 1$, this reestablishes Eq. (4) (to the second power of $\frac{1}{x}$).

## 3  A Reestimation Formula for Zipf's Law

Zipf's law concerns the asymptotic behavior of the relative frequencies $f(r)$ of a population as a function of rank $r$. It states that, asymptotically, the relative frequency is inversely proportional to rank:

$$f(r) = \frac{A}{B+r} \qquad (8)$$

This implies a finite total population, since the cumulative (i.e, the sum or integral) of the relative frequency over rank does not converge as rank approaches infinity:

$$F(r) = \begin{cases} \sum_{k=1}^r f(k) & \text{in the discrete case} \\ \int_1^r f(\rho)\, d\rho & \text{in the continuous case} \end{cases}$$

$$\lim_{r\to\infty} F(r) = \lim_{r\to\infty} A\ln(B+r) = \infty$$

To localize Zipf's law, we utilize Eq. (2) and observe that $r(x) = \frac{A}{f_x} - B = \frac{A'}{x} - B$,

$$\frac{N_{x+1}}{N_x} = \frac{r(x+1) - r(x+2)}{r(x) - r(x+1)} = \frac{\frac{A'}{x+1} - \frac{A'}{x+2}}{\frac{A'}{x} - \frac{A'}{x+1}} = \frac{\frac{1}{(x+1)\cdot(x+2)}}{\frac{1}{x\cdot(x+1)}} = \frac{x}{x+2} \qquad (9)$$

This suggests "Zipf's local reestimation formula"

$$x^* = (x+2)\cdot\frac{N_{x+1}}{N_x} \qquad (10)$$

which is deceptively similar to Turing's formula, Eq. (1), the only difference being that it assigns $\frac{x+2}{x+1}$ more relative-frequency mass to frequency count $x$.

### 3.1  Rederiving Zipf's law

If we rederive the asymptotic behavior, we again obtain Zipf's law. Assuming the recurrence equation

$$N_{x+1} = \frac{x}{x+2}\cdot N_x$$

we have that

$$N_{x+1} = \frac{x}{x+2} \cdot N_x = \frac{x \cdot \ldots \cdot 1}{(x+2) \cdot \ldots \cdot 3} \cdot N_1 = \frac{2}{(x+2) \cdot (x+1)} \cdot N_1 \approx \frac{C}{(x+1)^2} \quad (11)$$

We again use the equation for the derivative of the rank, Eq. (3), but now

$$\frac{dr}{df} = -N(x) \cdot N \approx -\frac{C}{x^2} \cdot N = -\frac{C'}{f^2}$$

Integration yields $r(f) = \dfrac{C'}{f} + C''$ and function inversion

$$f(r) = \frac{C'}{r - C''}$$

Identifying $C'$ with $A$ and $C''$ with $-B$ recovers Eq. (8).

## 4  A General Correspondence

If we generalize the rederivation of Zipf's law in Eq. (11) to $p = 2, 3, \ldots, x$, we find that

$$N_{x+1} = \frac{x}{x+p} \cdot N_x = \frac{x \cdot \ldots \cdot 1}{(x+p) \cdot \ldots \cdot (1+p)} \cdot N_1 = \frac{x!}{\prod_{k=1}^{x}(k+p)} \cdot N_1 \approx \frac{C}{(x+1)^p}$$

We integrate $\dfrac{C'}{f^p}$ to get $r(f) = \dfrac{C'}{f^{p-1}} + C''$, yielding a $r^{-\frac{1}{p-1}}$ asymptote for $f(r)$.

Although a nontrivial generalization, it is in fact the case that for real-valued $\theta : 1 \neq \theta \leq x$,

$$N_{x+1} = \frac{x}{x+\theta} \cdot N_x \quad (12)$$

results in the asymptote[1]

$$f(r) = C r^{-\frac{1}{\theta-1}} \quad (13)$$

The key observation here is that also for real-valued $\theta \leq x$ in general,

$$\frac{x!}{\prod_{k=1}^{x}(k+\theta)} \approx \frac{C}{(x+1)^{\theta}}$$

This means that we have a single reestimation equation

$$x^* = (x+\theta) \cdot \frac{N_{x+1}}{N_x} \quad (14)$$

parameterized by the real-valued parameter $\theta$, with the asymptotic behavior

$$f(r) = \begin{cases} C r^{-\frac{1}{\theta-1}} & \theta \neq 1 \\ C e^{-\lambda r} & \theta = 1 \end{cases} \quad (15)$$

Although this correspondence was derived with the requirement that $\theta \leq x$, we can in view of the discussion in Section 2.1 assume that $x$ is not only considerably larger than 1, but also greater than any fixed value of $\theta$. The extension to the negative real numbers is straightforward, although perhaps not very sensible. In fact, the convergence region for the cumulative of the frequency function as rank goes to infinity,

$$\sum_{r=1}^{\infty} f(r) \quad \text{or} \quad \int_{1}^{\infty} f(r)\, dr$$

is $\theta \in [1, 2)$, establishing Turing's formula and Zipf's law as the two extremes of this reestimation formula, in terms of resulting in a proper probability distribution for infinite populations; while the former does so, the latter does not.

---
[1] If $\theta = 1$, we have the Turing case with an exponentially declining asymptote, cf. Eq. (7).

## 4.1 Reversing the directions

Finally, assuming the asymptotic behavior of Eq. (13), we rederive the recurrence equation (12). The mathematics are very similar to those used to rederive Turing's formula in Section 2.3. Inverting the asymptotic behavior $f(r) = Cr^{-\frac{1}{\theta-1}}$ gives us $r(f) = \frac{C'}{f^{\theta-1}}$, which in turn yields

$$r(x) = \frac{C''}{x^{\theta-1}}$$

For notational convenience, let $\alpha$ denote $\theta - 1$, and assume that $0 < \theta \neq 1$, i.e., $-1 < \alpha \neq 0$.

$$\left| \frac{x}{x+\alpha+1} - \frac{N_{x+1}}{N_x} \right| =$$

$$= \left| \frac{x}{x+\alpha+1} - \frac{r(x+1) - r(x+2)}{r(x) - r(x+1)} \right| = \left| \frac{x}{x+\alpha+1} - \frac{\frac{1}{(x+1)^\alpha} - \frac{1}{(x+2)^\alpha}}{\frac{1}{x^\alpha} - \frac{1}{(x+1)^\alpha}} \right| =$$

$$= \left| \frac{x\left(\frac{1}{x^\alpha} - \frac{1}{(x+1)^\alpha}\right) - (x+\alpha+1)\left(\frac{1}{(x+1)^\alpha} - \frac{1}{(x+2)^\alpha}\right)}{(x+\alpha+1)\left(\frac{1}{x^\alpha} - \frac{1}{(x+1)^\alpha}\right)} \right| =$$

$$= \left| \frac{\left(\frac{x+1+\alpha}{(x+2)^\alpha} - \frac{x+1}{(x+1)^\alpha}\right) - \left(\frac{x+\alpha}{(x+1)^\alpha} - \frac{x}{x^\alpha}\right)}{(x+\alpha+1)\left(\frac{1}{x^\alpha} - \frac{1}{(x+1)^\alpha}\right)} \right| =$$

As before, the numerator can be written as $g(x+1) - g(x)$, now for $g(y) = \frac{y+\alpha}{(y+1)^\alpha} - \frac{1}{y^{\alpha-1}}$:

$$= \left| \frac{\int_x^{x+1} \left(\frac{\alpha-1}{y^\alpha} - \frac{(\alpha-1)y + \alpha^2 - 1}{(y+1)^{\alpha+1}}\right) dy}{(x+\alpha+1)\left(\frac{1}{x^\alpha} - \frac{1}{(x+1)^\alpha}\right)} \right| =$$

$$= \frac{|\alpha-1|}{x+\alpha+1} \cdot \left| \frac{\int_x^{x+1} \left(\frac{1}{y^\alpha} - \frac{1}{(y+1)^\alpha} - \frac{\alpha}{(y+1)^{\alpha+1}}\right) dy}{\frac{1}{x^\alpha} - \frac{1}{(x+1)^\alpha}} \right| =$$

$$= \frac{|\alpha-1|}{x+\alpha+1} \cdot \left| \frac{\int_x^{x+1} \left(\int_y^{y+1} \frac{\alpha}{z^{\alpha+1}} dz - \frac{\alpha}{(y+1)^{\alpha+1}}\right) dy}{\int_x^{x+1} \frac{\alpha}{y^{\alpha+1}} dy} \right| \leq$$

$$\leq \frac{|\alpha-1|}{x+\alpha+1} \cdot \frac{\int_x^{x+1} \left(\frac{1}{y^{\alpha+1}} - \frac{1}{(y+1)^{\alpha+1}}\right) dy}{\int_x^{x+1} \frac{1}{y^{\alpha+1}} dy} =$$

$$= \frac{|\alpha-1|}{x+\alpha+1} \cdot \frac{\int_x^{x+1} \left(\int_y^{y+1} \frac{\alpha+1}{z^{\alpha+2}} dz\right) dy}{\int_x^{x+1} \frac{1}{y^{\alpha+1}} dy} \leq$$

$$\leq \quad \frac{|\alpha^2-1|}{x+\alpha+1} \cdot \frac{\int_x^{x+1} \frac{1}{y^{\alpha+2}} dy}{\int_x^{x+1} \frac{1}{y^{\alpha+1}} dy} \quad \leq \quad \frac{|\alpha^2-1|}{x+\alpha+1} \cdot \frac{1}{x} \quad \leq \quad \frac{|\alpha^2-1|}{x^2}$$

This recaptures Eq. (12). Note that the derivation of Zipf's recurrence equation in Eq. (9) of Section 3 corresponds to the special case where $\alpha = 1$, i.e., where $\theta = 2$.

## 5  Conclusions

The relationship between Turing's formula and Zipf's law, which both concern population frequencies, was explored in the present article. The asymptotic behavior of the relative frequency as a function of rank implicit in one interpretation of Turing's local reestimation formula was derived and compared with Zipf's law. While the latter relates the rank and relative frequency as asymptotically inversely proportional, the former states that the frequency declines exponentially with rank. This means that while Zipf's law implies a finite total population, Turing's formula yields a proper probability distribution also for infinite populations.

In fact, it is tempting to interpret Turing's formula as smoothing the relative-frequency estimates towards a geometric distribution. This could potentially be used to improve sparse-data estimates by assuming a geometric distribution (tail), and introducing a ranking based on direct frequency counts, frequency counts when backing off to more general conditionings, order of appearance in the training data, or, to break any remaining ties, lexicographical order.

Conversely, a local reestimation formula in the vein of Turing's formula was derived from Zipf's law. Although the two equations are similar, Turing's formula shifts the frequency mass towards more frequent species. The two cases were generalized to a single spectrum of reestimation formulas and corresponding asymptotes, parameterized by one real-valued parameter. Furthermore, the two cases correspond to the upper and lower bounds of this parameter for which the cumulative of the frequency function converges as rank tends to infinity.

These results are in sharp contrast to common belief in the field; in [Baayen 1991], for example, we read: "Other models, such as Good (1953) ... have been put forward, all of which have Zipf's law as some special or limiting form." All of the Zipf-Simon-Mandelbrot distributions exhibit the same basic asymptotic behavior,

$$f(r) = \frac{C}{r^\beta}$$

parameterized by the positive real-valued parameter $\beta$. Comparing this with Eq. (15), we find that $\beta = \frac{1}{\theta-1} > 0$ and thus $\theta = 1 + \frac{1}{\beta} > 1$. In view of the established exponentially declining asymptote of the ideal Turing distribution, corresponding to $\theta = 1$, we can conclude that the latter is qualitatively different.

## Acknowledgements

This article originated from inspiring discussions with David Milward and Slava Katz. Many thanks! Most of the work was done while the author was visiting IRCS at the University of Pennsylvania at the invitation of Aravind Joshi, and a number of New York pubs at the invitation of Jussi Karlgren, both of which was very much appreciated. I wish to thank Mark Lauer for helpful comments and suggestions to improvements, Seif Haridi for constituting the entire audience at a seminar on this work and focusing the question session on the convergence region of the parameter $\theta$, and Åke Samuelsson for providing a bit of mathematical elegance.

I also gratefully acknowledge Rens Bod's encouraging comments and useful pointers to related work. Special credit is due to Mark Liberman for sharing his insights about Zipf's law, for drawing my attention to the Simon-Mandelbrot controversy, and for supplying various background material. This article, like others, has benefited greatly from comments by Khalil Sima'an.